\documentclass[10pt,twocolumn,letterpaper]{article}

\usepackage{cvpr}
\usepackage{times}
\usepackage{epsfig}
\usepackage{graphicx}
\usepackage{amsmath}
\usepackage{amssymb}
\usepackage{booktabs}
\usepackage{float}
\usepackage{caption}
\usepackage{subcaption}
\usepackage{alphalph}

\usepackage[breaklinks=true,bookmarks=false,colorlinks]{hyperref}

\cvprfinalcopy 

\def\cvprPaperID{****} 

\begin{document}

\title{Freeze the Discriminator: a Simple Baseline for Fine-Tuning GANs}

\author{
Sangwoo Mo\\
KAIST\\
{\tt\small swmo@kaist.ac.kr}
\and
Minsu Cho\\
POSTECH\\
{\tt\small mscho@postech.ac.kr}
\and
Jinwoo Shin\\
KAIST\\
{\tt\small jinwoos@kaist.ac.kr}
}

\maketitle

\begin{abstract}
Generative adversarial networks (GANs) have shown outstanding performance on a wide range of problems in computer vision, graphics, and machine learning, but often require numerous training data and heavy computational resources. To tackle this issue, several methods introduce a transfer learning technique in GAN training. They, however, are either prone to overfitting or limited to learning small distribution shifts. In this paper, we show that simple fine-tuning of GANs with frozen lower layers of the discriminator performs surprisingly well. This simple baseline, {\em FreezeD}, significantly outperforms previous techniques used in both unconditional and conditional GANs. We demonstrate the consistent effect using StyleGAN and SNGAN-projection architectures on several datasets of Animal Face, Anime Face, Oxford Flower, CUB-200-2011, and Caltech-256 datasets. The code and results are available at \url{https://github.com/sangwoomo/FreezeD}.
\end{abstract}


\section{Introduction}

Generative adversarial networks (GANs) \cite{goodfellow2014generative} have shown a remarkable success across a broad range of applications in computer vision, graphics, and machine learning, \eg, image generation \cite{brock2019large, karras2019style, karras2019analyzing}, image-to-image translation \cite{mo2019instagan, park2019semantic, choi2019stargan}, and video-to-video synthesis \cite{wang2018video, bansal2018recycle, chan2019everybody}. Current state-of-the-art GANs, however, often require a large amount of training data and heavy computational resources, which thus limits the applicability of GANs in practical scenarios. Numerous techniques have been proposed to overcome this limitation, \eg, transferring knowledge of a well-trained source model \cite{wang2018transferring, noguchi2019image, wang2019minegan}, learning meta-knowledge for quick adaptation to a target domain \cite{liu2019few, zakharov2019few, wang2019few}, using an auxiliary task to facilitate training \cite{chen2019self, lucic2019high, zhang2020consistency, zhao2020improved}, improving an inference procedure of suboptimal models \cite{azadi2019discriminator, turner2019metropolis, mo2019mining, tanaka2019discriminator}, using an expressive prior distribution \cite{gurumurthy2017deligan}, actively choosing samples to give supervision for conditional generation \cite{mo2019mining}, or actively sampling mini-batches for training \cite{sinha2019small}.

Among the approaches, transfer learning \cite{yosinski2014transferable} is arguably the most promising way to training models under limited data and resources. Indeed, most of the recent success in deep learning is built upon strong backbones pre-trained on large datasets in supervised \cite{deng2009imagenet} or self-supervised \cite{devlin2019bbert, he2019momentum} ways. Following the success of transferring classifiers in recognition tasks, one can also consider utilizing well-trained GAN backbones for downstream generation tasks. While several methods propose such transfer-learning approaches to training GANs \cite{wang2018transferring, noguchi2019image, wang2019minegan}, they are often prone to overfitting with limited training data \cite{wang2018transferring} or not robust in learning a significant distribution shift \cite{noguchi2019image, wang2019minegan}.

\begin{figure}
\centering
\includegraphics[width=.45\textwidth]{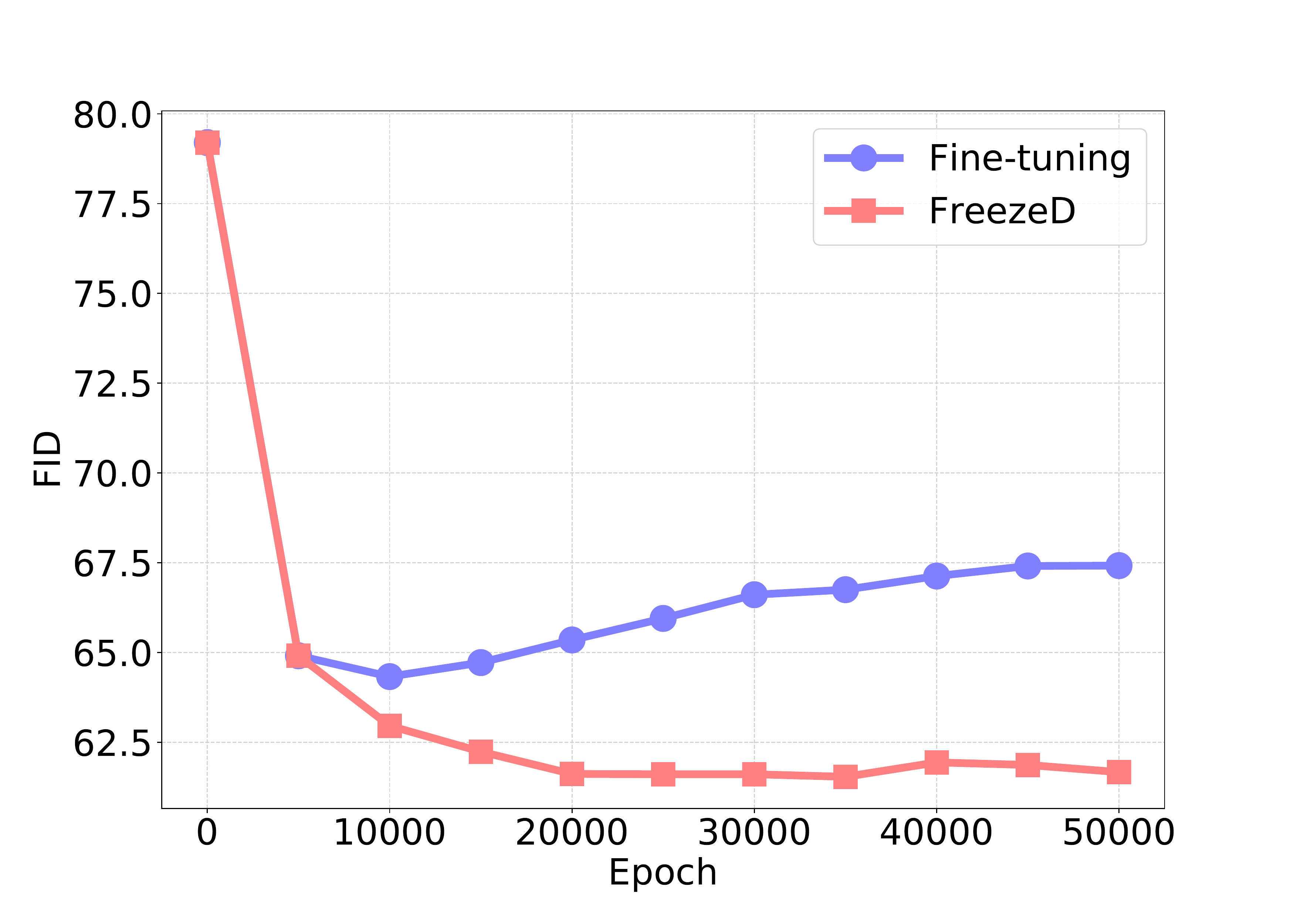}
\vspace{-0.1in}
\caption{
Trends of FID \cite{heusel2017gans} scores of fine-tuning and our proposed baseline, FreezeD, on `Dog' class in the Animal Face \cite{si2011learning} dataset. While fine-tuning suffers from overfitting, FreezeD shows consistent stability in training GANs.
} \label{fig:fig_trend}
\vspace{-0.1in}
\end{figure}

In this paper, we propose a simple yet effective baseline for transfer learning of GANs. In particular, we show that simple fine-tuning of GANs (both generator and discriminator) with frozen lower layers of the discriminator performs surprisingly well (see Figure \ref{fig:fig_trend}). Intuitively, the lower layers of the discriminator learn generic features of images while the upper layers learn to classify whether the image is real or fake based on the extracted features. We remark that this dichotomous view of a feature extractor and a classifier (and freezing the feature extractor for fine-tuning) is not new; it has been widely used for training classifiers \cite{yosinski2014transferable}. We confirm that this view is \textit{also} useful for GANs, and set its proper baseline for transfer learning of GANs.

We demonstrate the effectiveness of the simple baseline, dubbed {\em FreezeD}, using various architectures and datasets. For unconditional GANs, we fine-tune the StyleGAN \cite{karras2019style} architecture, which is pre-trained on FFHQ \cite{karras2019style}, onto Animal Face \cite{si2011learning} and Anime Face \cite{animeface} datasets, and for conditional GANs, fine-tune the SNGAN-projection \cite{miyato2018cgans} architecture, which is pre-trained on ImageNet \cite{deng2009imagenet}, onto Oxford Flower \cite{nilsback2008automated}, CUB-200-2011 \cite{wah2011caltech}, and Caltech-256 \cite{griffin2007caltech} datasets. FreezeD outperforms previous techniques for all experiment settings, \eg, improving the FID \cite{heusel2017gans} score from 64.28 of fine-tuning to 61.46 (-4.4\%) on `Dog' class of Animal Face dataset.

\section{Methods}

The goal of GANs \cite{goodfellow2014generative} is to learn a generator (and a corresponding discriminator) to match with a target data distribution. In transfer learning, we assume one can utilize a pre-trained source generator (and a corresponding discriminator) trained on the source data distribution to improve the target generator. See \cite{lucic2018gans, kurach2019large} for the survey of GANs. 

We first briefly review previous methods for transfer learning of GANs.
\begin{itemize}
    \item Fine-tuning \cite{wang2018transferring}: The most intuitive and effective way to transferring knowledge is fine-tuning; initialize the parameters of target models as the pre-trained weights of the source models. The authors report that fine-tuning \textit{both} the generator and the discriminator indeed shows the best performance.\footnote{It is more crucial for our case, as we use stronger source models.} However, fine-tuning often suffer from overfitting; hence one needs a proper regularization.

    \item Scale/shift \cite{noguchi2019image}: Since na\"ive fine-tuning is prone to overfitting, scale/shift suggest to update the normalization layers only (\eg, batch normalization (BN) \cite{ioffe2015batch}) while fixing all other weights. However, it often shows inferior results due to its restriction, especially when there is a significant shift between the source and the target distribution.
    
    \item Generative latent optimization (GLO) \cite{noguchi2019image, bojanowski2018optimizing}: Since GAN loss is given by the discriminator, which can be unreliable for limited data, GLO suggests fine-tuning the generator with supervised learning, where the loss is given by the sum of the L1 loss and the perceptual loss \cite{johnson2016perceptual}. Here, GLO jointly optimizes the generator and the latent codes to avoid overfitting; one latent code (and its corresponding generated sample) matches one real sample; hence, the generator can generalize samples by interpolation. While GLO improves the stability, it tends to produce blurry images due to the lack of adversarial loss (and prior knowledge of the source discriminator).

    \item MineGAN \cite{wang2019minegan}: To avoid overfitting of the generator, MineGAN suggests to fix the generator and modify the latent codes. To this end, MineGAN train a miner network that transforms the latent code to another latent code. While this importance-sampling-like approach can be effective when the source distribution and the target distributions share support, it may not be generalized when their supports are disjointed.
\end{itemize}

We now introduce a simple baseline, FreezeD, which outperforms the previous methods despite its simplicity, and suggest two other methods for possible future directions, which may give further improvement. We remark that our goal is not to advocate the state-of-the-art but to set a simple and effective baseline. By doing so, we hope to encourage new techniques that outperform the proposed baseline.
\begin{itemize}
    \item FreezeD (our proposed baseline): We find that simply freezing the lower layers of the discriminator and only fine-tune the upper layers performs surprisingly well. We call this simple yet effective baseline as {\em FreezeD}, and will demonstrate its consistent gain over the previous methods in the experimental section. 

    \item L2-SP \cite{li2018explicit}: In addition to the prior methods, we test L2-SP, which is known to be effective for the classifiers. Built upon to the fine-tuning, L2-SP regularizes the target models not to move far from the source models. In particular, it regularizes the L2-norm of the parameters of source models and target models. In our experiments, we applied L2-SP to the generator, discriminator, and both, but the results were not satisfactory. However, since freezing layers can be viewed as giving the infinite weight of L2-SP for the chosen layers and 0 for the other layers, using proper weights for each layer may perform better.

    \item Feature distillation \cite{hinton2014distilling, romero2015fitnets}: We also test feature distillation, one of the most popular approaches to transfer learning of classifiers. Among the variants, we simply distill the activations of the source models and target models (initialized to the source models). We find that feature distillation shows comparable results to FreezeD while takes twice computation. Investigating more advanced techniques (\eg, \cite{ahn2019variational, jang2019learning, park2019relational}) would be an interesting and promising future direction.\footnote{We observe that feature distillation shows more stable (but similar best) results than FreezeD for SNGAN-projection experiments.}
\end{itemize}

\section{Experiments}

\begin{figure*}[t]
\centering
\begin{subfigure}{0.32\textwidth}
\includegraphics[width=\textwidth]{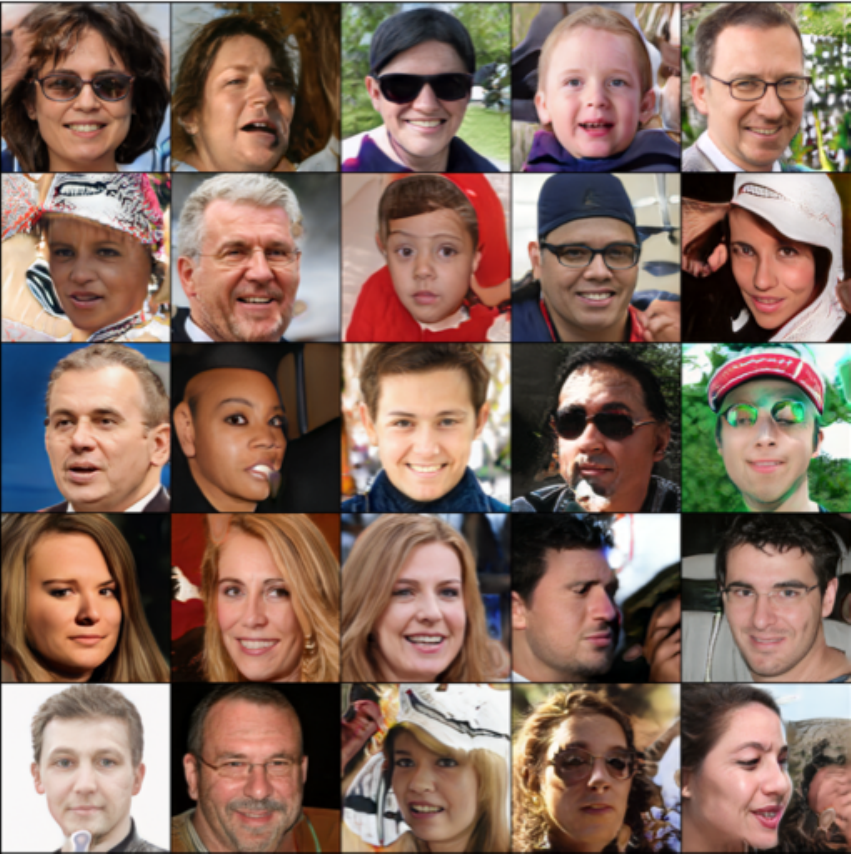}
\caption{Original (FFHQ)}
\end{subfigure}~
\begin{subfigure}{0.32\textwidth}
\includegraphics[width=\textwidth]{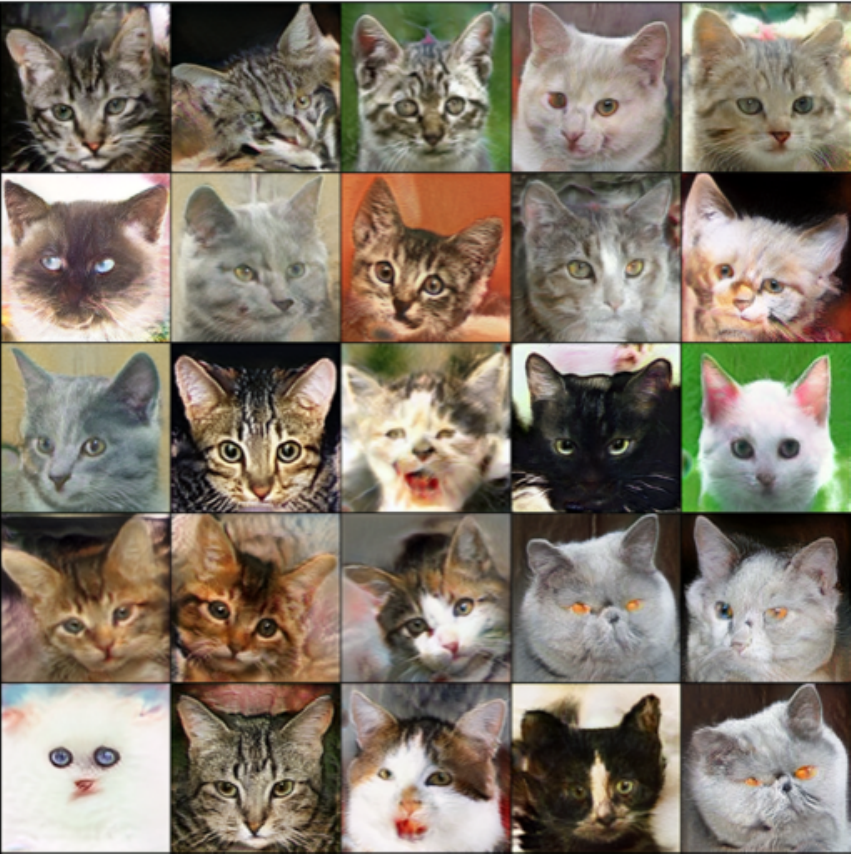}
\caption{Cat}
\end{subfigure}~
\begin{subfigure}{0.32\textwidth}
\includegraphics[width=\textwidth]{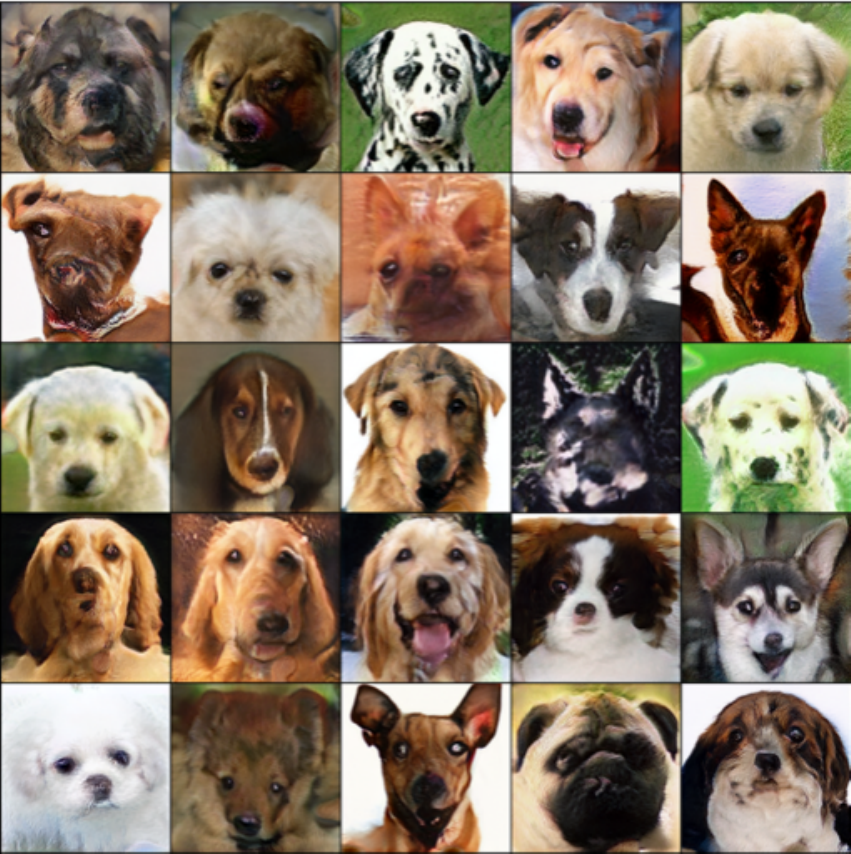}
\caption{Dog}
\end{subfigure}
\vspace{-0.05in}
\caption{Samples generated by StyleGAN of (a) original weights, and trained by FreezeD under (b) `Cat', and (c) `Dog' classes in the Animal Face dataset. Each entry indicates the same latent code. Same latent code shares the same semantics even after fine-tuning, \eg, the background color and hair color are preserved. See Appendix \ref{sec:a_samples_stylegan} for more qualitative results.} \label{fig:samples_stylegan}
\vspace{-0.05in}
\end{figure*}

\begin{table*}
\caption{FID scores under  Animal Face dataset. Left and right values indicate the best and final FID scores.} \label{tab:fid_animal}
\vspace{-0.15in}
\resizebox{\textwidth}{!}{%
\begin{tabular}{lcccccccccc}
	\toprule
	            & Bear & Cat & Chicken & Cow & Deer & Dog & Duck & Eagle & Elephant & Human\\
	\midrule
	Fine-tuning & 82.82/84.38 & 71.76/73.47 & 88.10/88.83 & 87.07/87.46 & 82.11/84.04 & 64.28/67.42 & 92.54/92.54 & 85.52/86.88 & 84.10/84.33 & 76.62/76.72\\
	FreezeD & \textbf{78.77}/\textbf{78.77} & \textbf{69.64}/\textbf{69.97} & \textbf{86.20}/\textbf{86.53} & \textbf{84.32}/\textbf{84.39} & \textbf{78.67}/\textbf{79.73} & \textbf{61.46}/\textbf{61.67} & \textbf{88.82}/\textbf{89.14} & \textbf{82.15}/\textbf{82.62} & \textbf{80.00}/\textbf{80.24} & \textbf{73.51}/\textbf{73.89}\\
	\midrule
	            & Lion & Monkey & Mouse & Panda & Pigeon & Pig & Rabbit & Sheep & Tiger & Wolf\\
	\midrule
	Fine-tuning & 76.86/78.36 & 86.70/87.30 & 84.95/85.61 & 74.29/76.07 & 81.24/81.36 & 85.31/86.08 & 89.11/89.82 & 86.98/87.89 & 73.21/75.06 & 79.97/81.37\\
	FreezeD & \textbf{73.49}/\textbf{73.59} & \textbf{82.31}/\textbf{82.61} & \textbf{81.72}/\textbf{82.30} & \textbf{72.19}/\textbf{72.62} & \textbf{77.79}/\textbf{78.07} & \textbf{83.22}/\textbf{83.31} & \textbf{85.65}/\textbf{85.65} & \textbf{84.33}/\textbf{84.55} & \textbf{71.26}/\textbf{71.54} & \textbf{76.47}/\textbf{76.47}\\
	\bottomrule
\end{tabular}}
\vspace{0.05in}
\caption{FID scores under the Anime Face dataset. Left and right values indicate the best and final FID scores.} \label{tab:fid_anime}
\vspace{-0.15in}
\resizebox{\textwidth}{!}{%
\begin{tabular}{lcccccccccc}
	\toprule
	            & Miku & Sakura & Haruhi & Fate & Nanoha & Lelouch & Mio & Yuki & Shana & Reimu\\
	\midrule
	Fine-tuning & 95.54/98.44 & 66.94/67.43 & 76.34/77.44 & 79.81/83.94 & 71.03/72.04 & 83.58/84.11 & 86.14/88.24 & 81.38/83.12 & 79.05/79.79 & 80.82/82.44\\
	FreezeD & \textbf{93.37}/\textbf{95.63} & \textbf{65.40}/\textbf{65.91} & \textbf{74.50}/\textbf{74.56} & \textbf{77.76}/\textbf{78.80} & \textbf{68.41}/\textbf{68.41} & \textbf{80.20}/\textbf{82.31} & \textbf{81.55}/\textbf{85.90} & \textbf{79.65}/\textbf{79.83} & \textbf{77.39}/\textbf{77.39} & \textbf{79.27}/\textbf{79.31}\\
	\bottomrule
\end{tabular}}
\vspace{0.05in}
\caption{Comparison of various methods under `Cat' and `Dog' classes in the Animal Face dataset. Left and right values indicate the best and final FID scores. $^\dagger$ indicates the model is trained by GLO loss, otherwise by GAN loss.} \label{tab:fid_comparison}
\vspace{-0.15in}
\resizebox{\textwidth}{!}{%
\begin{tabular}{lcccccccccc}
	\toprule
	    & Fine-tuning & Fine-tuning$^\dagger$ & Scale/shift & Scale/shift$^\dagger$ & MineGAN & MineGAN$^\dagger$ & L2-SP ($G$) & L2-SP ($D$) & L2-SP ($G$,$D$) & FreezeD \\
	\midrule
	Cat & 71.76/73.47 & 78.21/78.32 & 71.99/73.42 & 80.63/80.63 & 82.67/82.67 & 82.68/82.95 & 71.77/73.78 & 71.54/72.67 & 71.70/73.47 & \textbf{69.64}/\textbf{69.97}\\
	Dog & 64.28/67.42 & 75.19/75.45 & 64.12/67.79 & 79.08/79.91 & 79.05/79.23 & 79.11/79.20 & 64.18/67.14 & 64.28/66.68 & 64.25/66.06 & \textbf{61.46}/\textbf{61.67}\\
	\bottomrule
\end{tabular}}
\end{table*}

\begin{figure*}[t]
\centering
\begin{subfigure}{0.49\textwidth}
\includegraphics[width=\textwidth]{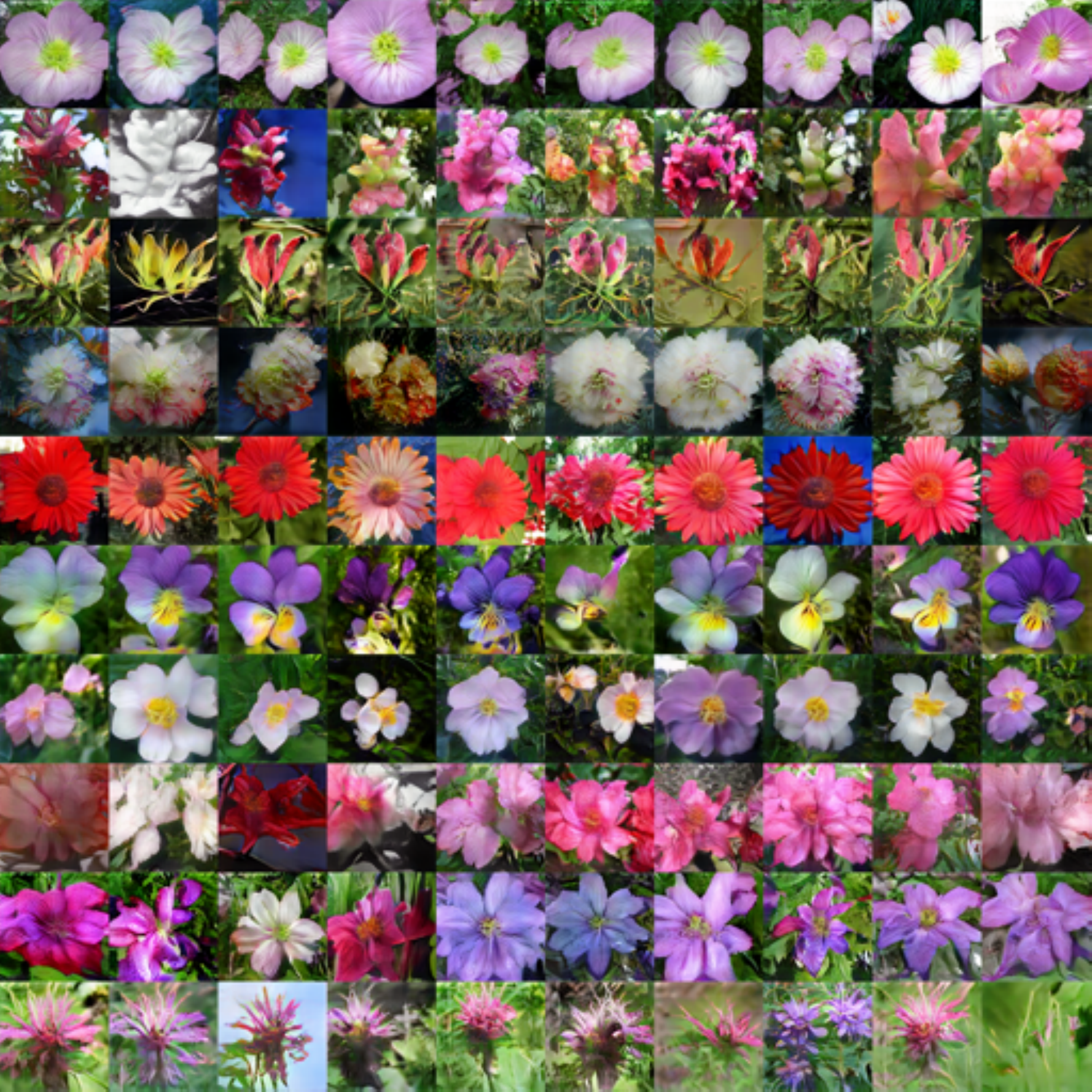}
\caption{Flower (fine-tuning)}
\end{subfigure}~
\begin{subfigure}{0.49\textwidth}
\includegraphics[width=\textwidth]{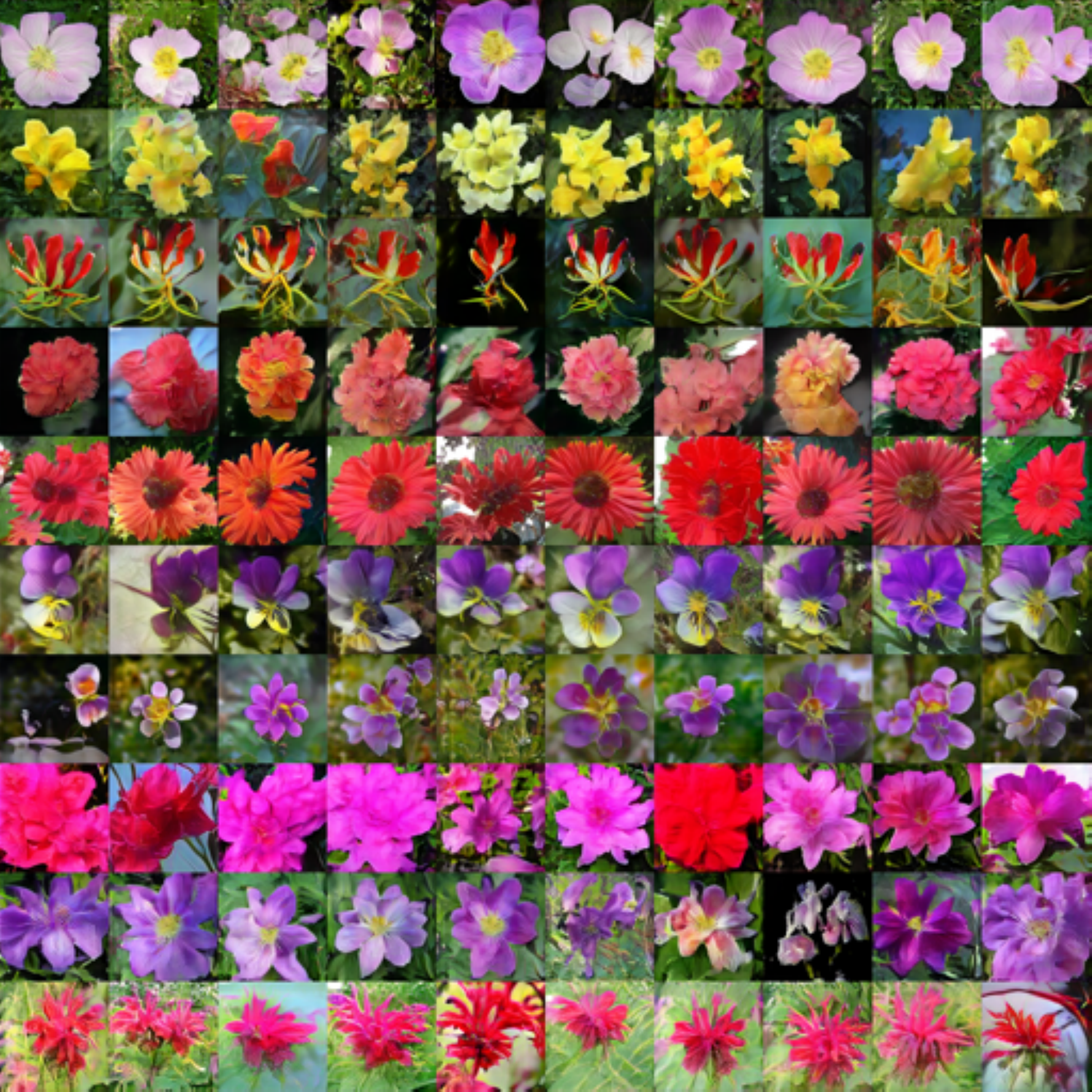}
\caption{Flower (FreezeD)}
\end{subfigure}~
\vspace{-0.05in}
\caption{Samples generated by SNGAN-projection trained by (a) fine-tuning and (b) FreezeD under the Oxford Flower dataset. Each row indicates the same class. FreezeD generates more class-consistent samples than fine-tuning, \eg, fine-tuning generates some abnormal samples for row 2 and 8. See Appendix \ref{sec:a_samples_projection} for more qualitative results.} \label{fig:sample_projection}
\vspace{-0.1in}
\end{figure*}

In this section, we demonstrate the effectiveness of the simple yet effective baseline, FreezeD. We conduct extensive experiments for both unconditional GANs and conditional GANs in Section \ref{sec:exp_uncond} and Section \ref{sec:exp_cond}, respectively.

\subsection{Unconditional GAN}
\label{sec:exp_uncond}

We first demonstrate results for unconditional GANs. We use the StyleGAN \cite{karras2019style} architecture pre-trained on FFHQ \cite{karras2019style} dataset, and fine-tune it on Animal Face \cite{si2011learning} and Anime Face \cite{animeface} datasets. We use full 20 classes of the Animal Face dataset, and the first 10 classes among the total 1,000 classes of the Anime Face dataset. Each class contains around 100 samples. We use the public pre-trained model\footnote{\url{https://github.com/rosinality/style-based-gan-pytorch}} of resolution 256$\times$256 and fine-tune the models following the original training scheme for 50,000 iterations. We remark that the training performed successfully without progressive training by utilizing the source models.

Figure \ref{fig:samples_stylegan} visualizes the generated samples using the original weights and the fine-tuned weights on `Cat' and `Dog' classes in the Animal Face dataset. Notably, the same latent code shares the same semantics even after fine-tuning. See Appendix \ref{sec:a_samples_stylegan} for more qualitative results. We also evaluate the FID \cite{heusel2017gans} scores of the vanilla fine-tuning and FreezeD under Animal Face and Anime Face datasets in Table \ref{tab:fid_animal} and Table \ref{tab:fid_anime}, respectively. We freeze the discriminator until layer 4. See Appendix \ref{sec:a_layers} for the ablation study on different layers. FreezeD improves both the best performance and the stability as shown by the best and final FID scores.

We finally compare FreezeD with several previous methods, including scale/shift, GLO, MineGAN, L2-SP, and feature distillation (FD). We choose the weights of L2-SP and FD from $\{0.1,1,10\}$ and simply use $1$ for all experiments. We follow the hyperparameters of \cite{noguchi2019image} for GLO, and use 2-layer MLP with ReLU activation for the Miner network. Table \ref{tab:fid_comparison} presents the FID scores of each method. Feature distillation and qualitative results are in Appendix \ref{sec:a_fm} and \ref{sec:a_compare}, respectively. Scale/shift and L2-SP are too restrictive and thus harms diversity. GLO produces blurry images while MineGAN fails to learn the distribution shift.

\subsection{Conditional GAN}
\label{sec:exp_cond}

We also demonstrate the results for conditional GANs. We use the SNGAN-projection \cite{miyato2018cgans} architecture pre-trained on ImageNet \cite{deng2009imagenet} dataset, and fine-tune it on Oxford Flower \cite{nilsback2008automated}, CUB-200-2011 \cite{wah2011caltech}, and Caltech-256 \cite{griffin2007caltech} datasets. Each dataset contains 102, 200, and 256 classes, respectively, where each class has 50-100 samples. We use the public pre-trained model\footnote{\url{https://github.com/pfnet-research/sngan_projection}} of resolution 128$\times$128 and fine-tune the networks following the original training scheme for 20,000 iterations. SNGAN-projection has a larger variance than StyleGAN, but still the trend is similar.

Figure \ref{fig:sample_projection} visualizes the samples generated using the model trained by fine-tuning and FreezeD. FreezeD generates more class-consistent samples than fine-tuning as shown in the 2nd and 8th rows. See Appendix \ref{sec:a_samples_projection} for more qualitative results.
We also evaluate the FID \cite{heusel2017gans} scores of the vanilla fine-tuning and FreezeD in Table \ref{tab:fid_projection}. We freeze the discriminator until \{3, 2, 1\} layers for \{Oxford Flower, CUB-200-2011, Caltech-256 datasets\}, respectively, as the distribution shift goes larger. See Appendix \ref{sec:a_layers} for details. FreezeD improves both the performance and stability for most cases, but harms the stability for Oxford Flower. We find that feature distillation shows more stable results in our experiments. We leave this investigation for future work.

\begin{table}[t]
\centering
\caption{FID scores under SNGAN-projection architecture. Left and right values indicate the best and final FID scores.} \label{tab:fid_projection}
\vspace{-0.1in}
\resizebox{.47\textwidth}{!}{%
\begin{tabular}{lccc}
	\toprule
	            & Oxford Flower & CUB-200-2011 & Caltech-256\\
	\midrule
	Fine-tuning & 27.05/\textbf{32.51} & 32.29/32.60 & 62.20/63.37\\
	FreezeD & \textbf{24.80}/52.92 & \textbf{26.37}/\textbf{27.63} & \textbf{60.53}/\textbf{60.53}\\
	\bottomrule
\end{tabular}}
\vspace{-0.1in}
\end{table}
\section{Conclusion}

We have introduced a simple yet effective baseline, FreezeD, for transfer learning of GANs. FreezeD splits the discriminator into a feature extractor and a classifier and then fine-tune the classifier only. We demonstrate that this simple baseline clearly outperforms most of the previous methods using various architectures and datasets. Our observation raises two questions. First, the transferability of the feature extractor of the discriminator could be applied for the universal detector of generated images \cite{wang2019cnn}. Second, one can design a more sophisticated method that outperforms our proposed baseline. We hypothesize that the advanced version of feature distillation \cite{hinton2014distilling, romero2015fitnets} could be a promising direction.

\clearpage
{\small
\bibliographystyle{abbrv}

}

\clearpage
\appendix

\onecolumn
\begin{center}{\bf {\LARGE Supplementary Material:}}\end{center}
\begin{center}{\bf {\Large Freeze the Discriminator: a Simple Baseline for Fine-Tuning GANs}}\end{center}

\vspace{0.1in}
\section{Ablation Study on Freezing Layers}
\label{sec:a_layers}

We study the effect of freezing layers of the discriminator for StyleGAN and SNGAN-projection in Table \ref{tab:a_layer_stylegan} and Table \ref{tab:a_layer_projection}, respectively. In StyleGAN, layer 4 consistently shows the best performance. However, in SNGAN-projection, layer \{3, 2, 1\} were the best for Oxford Flower, CUB-200-2011, and Caltech-256 datasets, respectively. It is since Caltech-256 is harder to learn compared to Oxford Flower (\ie, distribution shift is larger). Intuitively, one should less restrict the model to adapt to the large distribution shift. One can also see that FreezeD is less stable than fine-tuning for the Oxford Flower dataset. We observe that feature distillation shows better stability while showing a similar best performance in our early experiments. Investigating a more sophisticated method would be an interesting research direction.

\begin{table*}[h]
\centering\small
\caption{Ablation study on freezing layers of $D$ on StyleGAN architecture under `Cat' and `Dog' classes in the Animal Face dataset. Layer $i$ indicates that the first $i$ layers of the discriminator are frozen. Layer 4 performs the best.} \label{tab:a_layer_stylegan}
\begin{tabular}{lcccccccc}
	\toprule
	    & Fine-tuning & Layer 1 & Layer 2 & Layer 3 & Layer 4 & Layer 5 & Layer 6 & Layer 7\\
	\midrule
	Cat & 71.76/73.47 & 71.44/72.57 & 70.81/71.86 & 70.03/70.46 & \textbf{69.64}/\textbf{69.97} & 70.12/70.40 & 70.93/75.92 & 79.41/85.26\\
	Dog & 64.28/67.42 & 63.63/66.69 & 63.18/64.88 & 61.85/62.65 & \textbf{61.46}/\textbf{61.67} & 62.42/62.86 & 63.51/64.15 & 76.52/87.86\\
	\bottomrule
\end{tabular}
\end{table*}

\begin{table*}[h]
\centering\small
\caption{Ablation study on freezing layers of $D$ on SNGAN-projection architecture under Oxford Flower, CUB-200-2011, Caltech-256 datasets. Layer $i$ indicates that the first $i$ layers of the discriminator are frozen.} \label{tab:a_layer_projection}
\begin{tabular}{lcccccc}
	\toprule
	    & Fine-tuning & Layer 1 & Layer 2 & Layer 3 & Layer 4 & Layer 5\\
	\midrule
	Oxford Flower & 27.05/\textbf{32.51} & 27.65/42.14 & 25.85/42.31 & \textbf{24.80}/52.92 & 25.41/87.60 & 25.35/104.07\\
	CUB-200-2011 & 32.29/32.60 & 28.80/31.80 & \textbf{26.37}/\textbf{27.63} & 28.48/28.48 & 26.87/29.29 & 29.92/34.08\\
	Calteth-256 & 62.20/63.37 & \textbf{60.53}/\textbf{60.53} & 61.59/61.94 & 61.29/61.95 & 61.92/62.88 & 62.90/62.90\\
	\bottomrule
\end{tabular}
\end{table*}

\clearpage
\section{Comparison to Feature Distillation}
\label{sec:a_fm}

We compare FreezeD with feature distillation. We linearize the activations of the $i$-th layer of the discriminator, and match the activations of the source and target discriminators. Since the activation has a different size for each layer, we use the L2-norm normalized by the feature dimension. We simply use $1$ for the weight of the regularizer regardless of the layer. Table \ref{tab:a_comparison} presents the comparison results. Feature distillation and FreezeD shows comparable results, while feature distillation is twice slower. Hence, we choose to FreezeD as the baseline for this paper.

\begin{table*}[h]
\centering\small
\caption{Comparison of FreezeD and feature distillation (FD) on StyleGAN architecture under `Bear', `Cat', and `Dog' classes in the Animal Face dataset. FM (layer $i$) indicates the activations after layer $i$ are matched. Feature distillation shows comparable results to FreezeD while it is twice slower.} \label{tab:a_comparison}
\begin{tabular}{lcccc}
	\toprule
	    & Fine-tuning & FreezeD & FD (layer 4) & FD (layer 5)\\
	\midrule
	Bear & 82.82/84.38 & \textbf{78.77}/\textbf{78.77} & 79.47/80.10 & 79.41/79.64\\
	Cat & 71.76/73.47 & 69.64/69.97 & 69.45/\textbf{69.75} & \textbf{69.35}/69.80\\
	Dog & 64.28/67.42 & 61.46/61.67 & 61.50/62.00 & \textbf{61.31}/\textbf{61.44}\\
	\bottomrule
\end{tabular}
\end{table*}

\vspace{0.1in}
\section{Qualitative Results for Prior Methods}
\label{sec:a_compare}

\begin{figure*}[h]
\centering
\begin{subfigure}{0.24\textwidth}
\includegraphics[width=\textwidth]{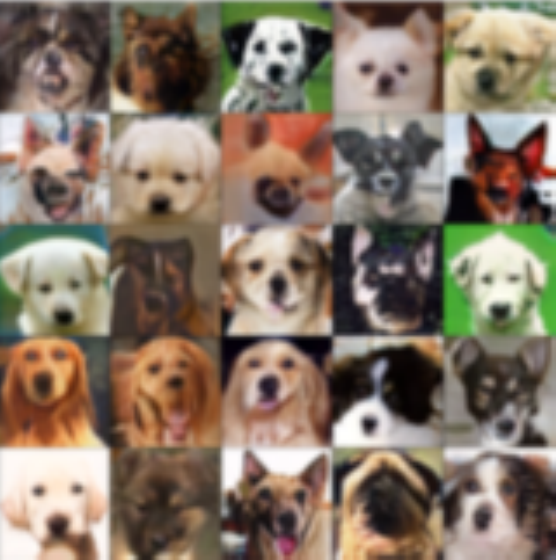}
\caption{Scale/shift}
\end{subfigure}~
\begin{subfigure}{0.24\textwidth}
\includegraphics[width=\textwidth]{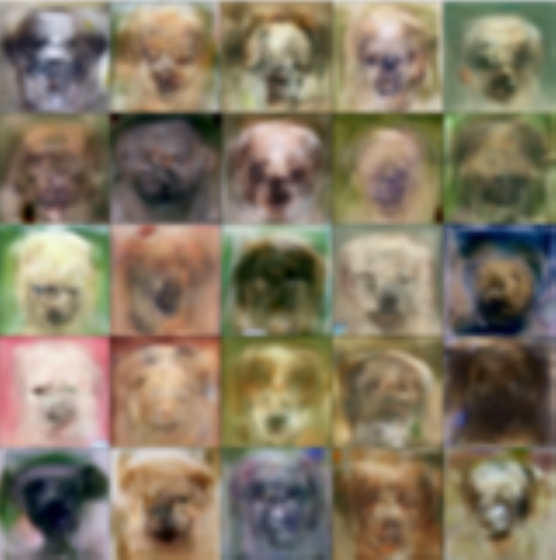}
\caption{GLO}
\end{subfigure}~
\begin{subfigure}{0.24\textwidth}
\includegraphics[width=\textwidth]{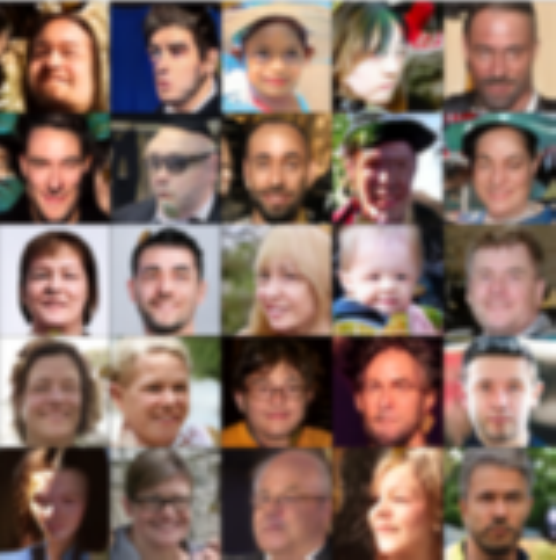}
\caption{MineGAN}
\end{subfigure}~
\begin{subfigure}{0.24\textwidth}
\includegraphics[width=\textwidth]{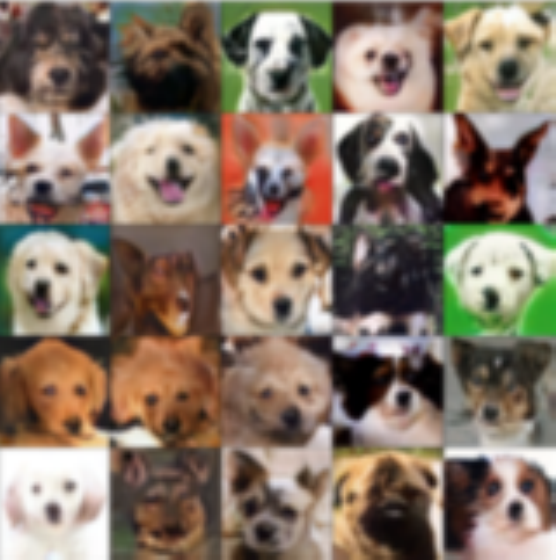}
\caption{L2-SP ($G$,$D$)}
\end{subfigure}
\vspace{-0.05in}
\caption{Samples generated by prior methods under `Dog' class in the Animal Face dataset.} \label{fig:a_compare}
\end{figure*}

We visualize the samples generated by the prior methods in Figure \ref{fig:a_compare}. Scale/shift and L2-SP generates reasonable samples, but have less diversity as measured by FID scores. GLO generates blurry images due to the lack of adversarial loss and the knowledge of source discriminator. In our experiments, MineGAN totally fails to adapt to the target distribution. Note that MineGAN assumes the source distribution covers (or at least close to) the target distribution (\eg, adult faces to child faces as in the original paper \cite{wang2019minegan}), but cannot be applied if the distributions have disjoint support (\eg, human faces to dog faces).

\clearpage
\section{Generated Samples by StyleGAN}
\label{sec:a_samples_stylegan}

\begin{figure}[H]
\centering
\begin{subfigure}{0.32\textwidth}
\includegraphics[width=\textwidth]{figures/original.pdf}
\caption{Original (FFHQ) \cite{karras2019style}}
\end{subfigure}~
\begin{subfigure}{0.32\textwidth}
\includegraphics[width=\textwidth]{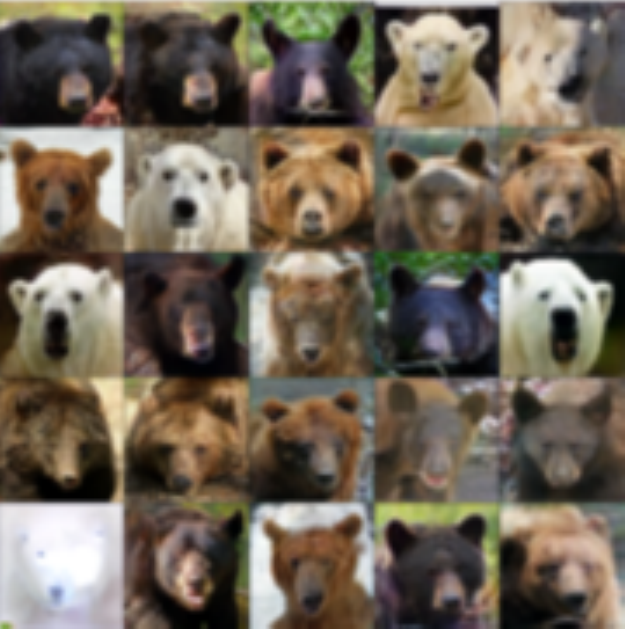}
\caption{Bear}
\end{subfigure}~
\begin{subfigure}{0.32\textwidth}
\includegraphics[width=\textwidth]{figures/cat.pdf}
\caption{Cat}
\end{subfigure}
\end{figure}

\begin{figure}[H]
\centering
\ContinuedFloat 
\begin{subfigure}{0.32\textwidth}
\includegraphics[width=\textwidth]{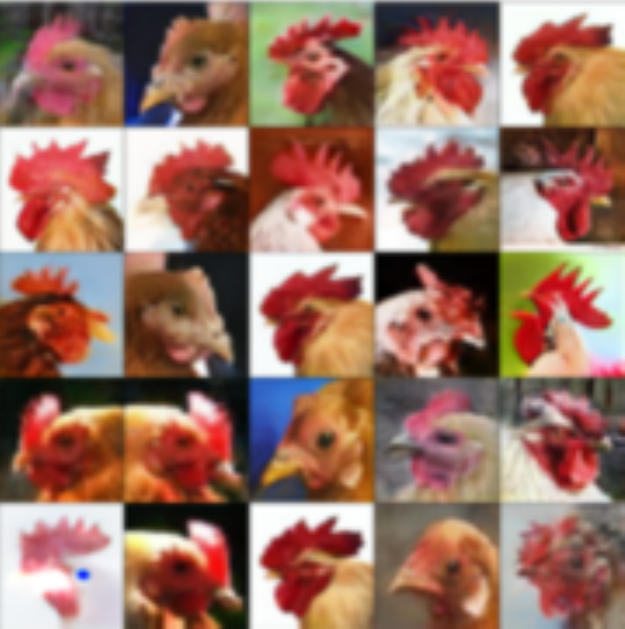}
\caption{Chicken}
\end{subfigure}~
\begin{subfigure}{0.32\textwidth}
\includegraphics[width=\textwidth]{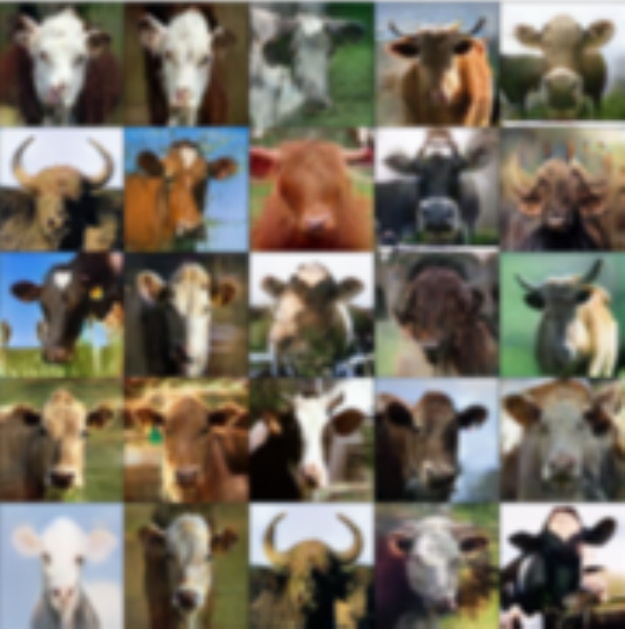}
\caption{Cow}
\end{subfigure}~
\begin{subfigure}{0.32\textwidth}
\includegraphics[width=\textwidth]{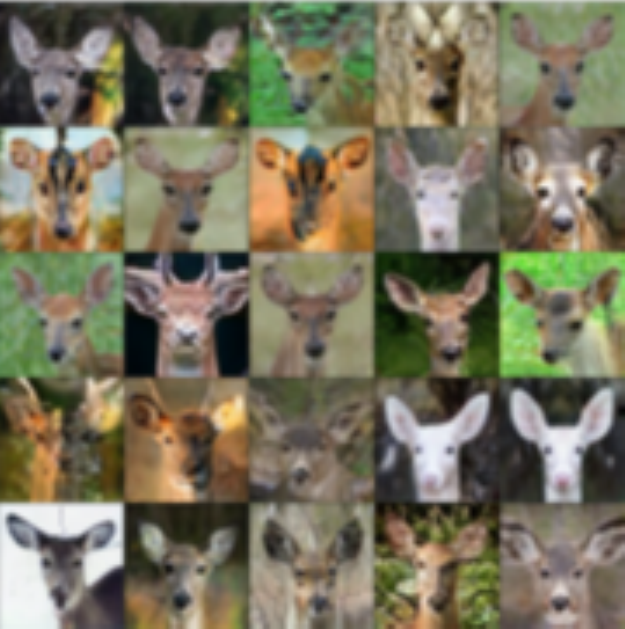}
\caption{Deer}
\end{subfigure}
\end{figure}

\begin{figure}[H]
\centering
\ContinuedFloat 
\begin{subfigure}{0.32\textwidth}
\includegraphics[width=\textwidth]{figures/dog.pdf}
\caption{Dog}
\end{subfigure}~
\begin{subfigure}{0.32\textwidth}
\includegraphics[width=\textwidth]{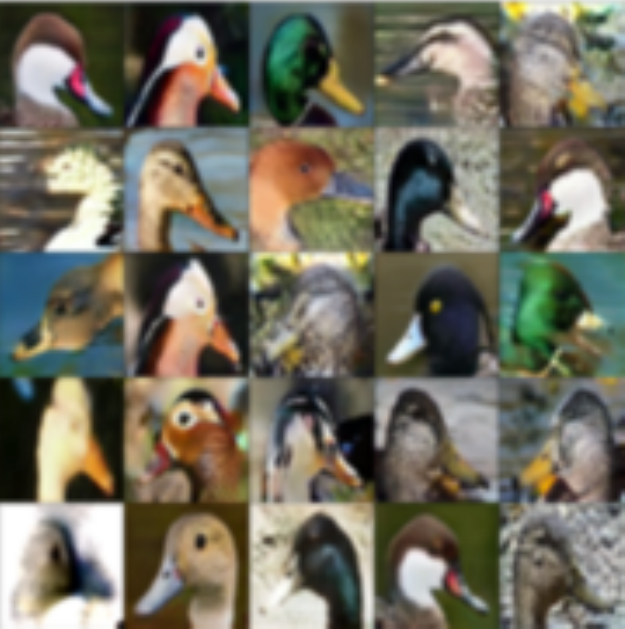}
\caption{Duck}
\end{subfigure}~
\begin{subfigure}{0.32\textwidth}
\includegraphics[width=\textwidth]{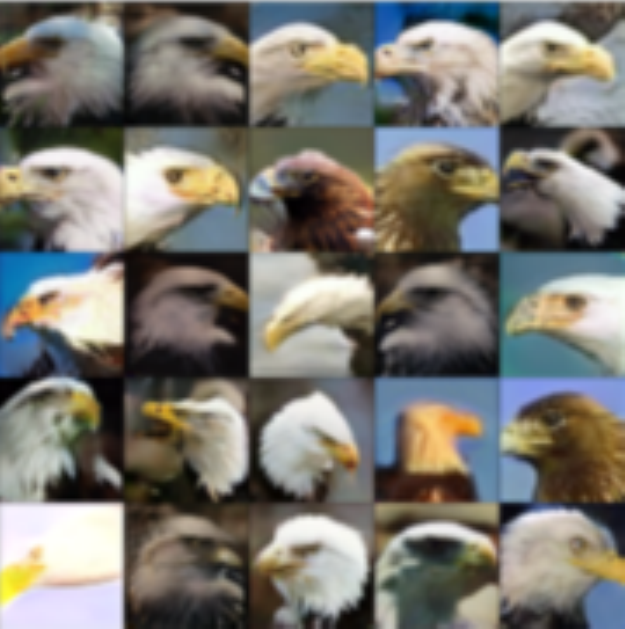}
\caption{Eagle}
\end{subfigure}
\end{figure}

\clearpage

\begin{figure}[H]
\centering
\ContinuedFloat 
\begin{subfigure}{0.32\textwidth}
\includegraphics[width=\textwidth]{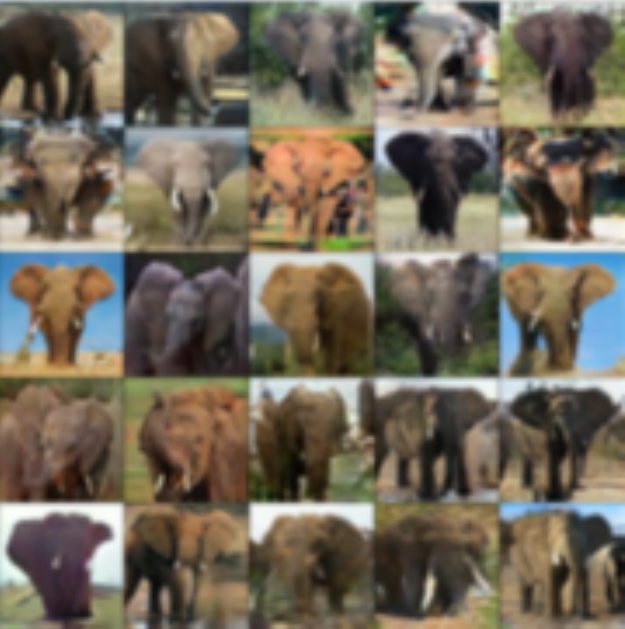}
\caption{Elephant}
\end{subfigure}~
\begin{subfigure}{0.32\textwidth}
\includegraphics[width=\textwidth]{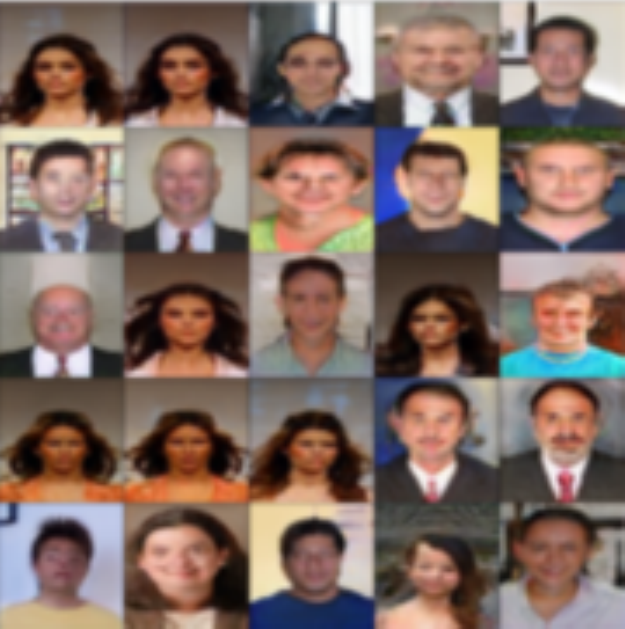}
\caption{Human}
\end{subfigure}~
\begin{subfigure}{0.32\textwidth}
\includegraphics[width=\textwidth]{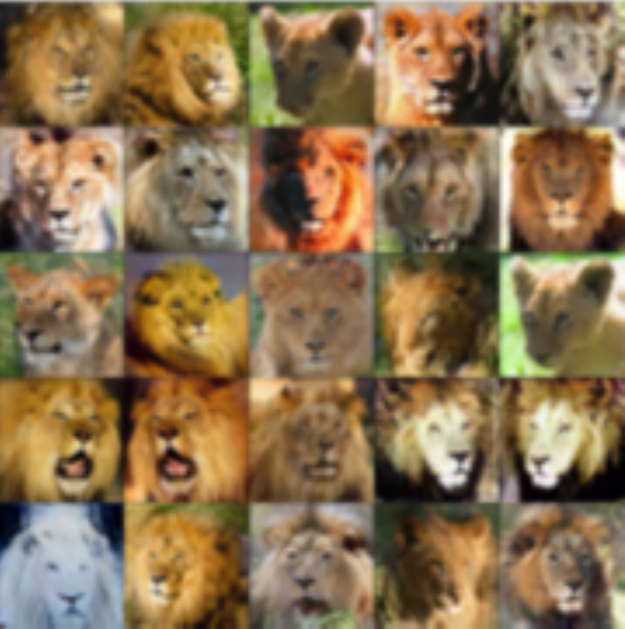}
\caption{Lion}
\end{subfigure}
\end{figure}

\begin{figure}[H]
\centering
\ContinuedFloat 
\begin{subfigure}{0.32\textwidth}
\includegraphics[width=\textwidth]{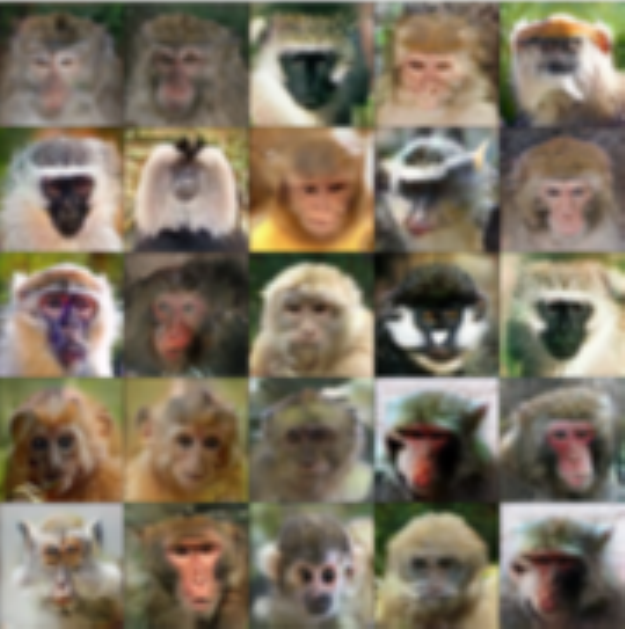}
\caption{Monkey}
\end{subfigure}~
\begin{subfigure}{0.32\textwidth}
\includegraphics[width=\textwidth]{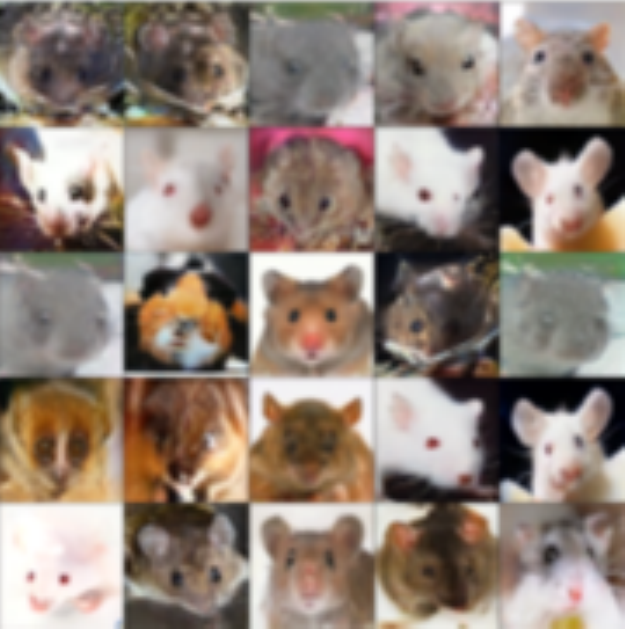}
\caption{Mouse}
\end{subfigure}~
\begin{subfigure}{0.32\textwidth}
\includegraphics[width=\textwidth]{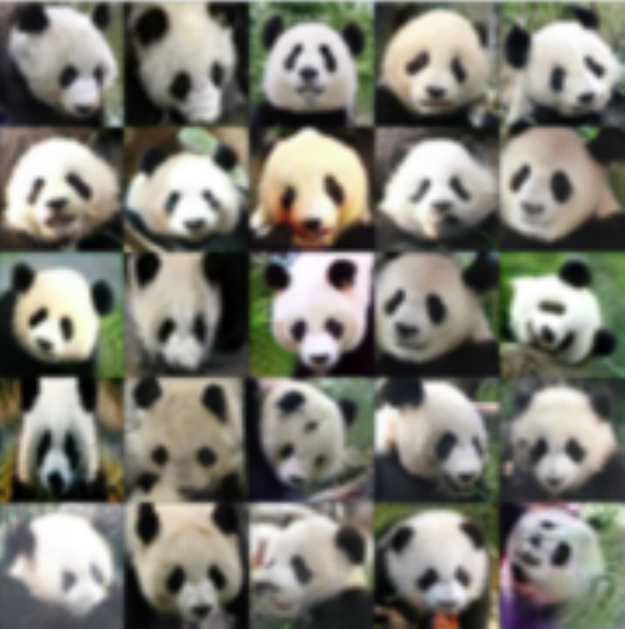}
\caption{Panda}
\end{subfigure}
\end{figure}

\begin{figure}[H]
\centering
\ContinuedFloat 
\begin{subfigure}{0.32\textwidth}
\includegraphics[width=\textwidth]{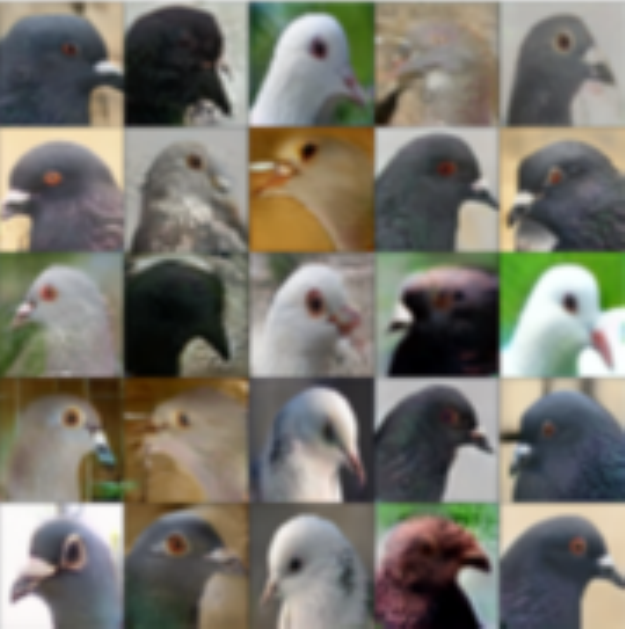}
\caption{Pigeon}
\end{subfigure}~
\begin{subfigure}{0.32\textwidth}
\includegraphics[width=\textwidth]{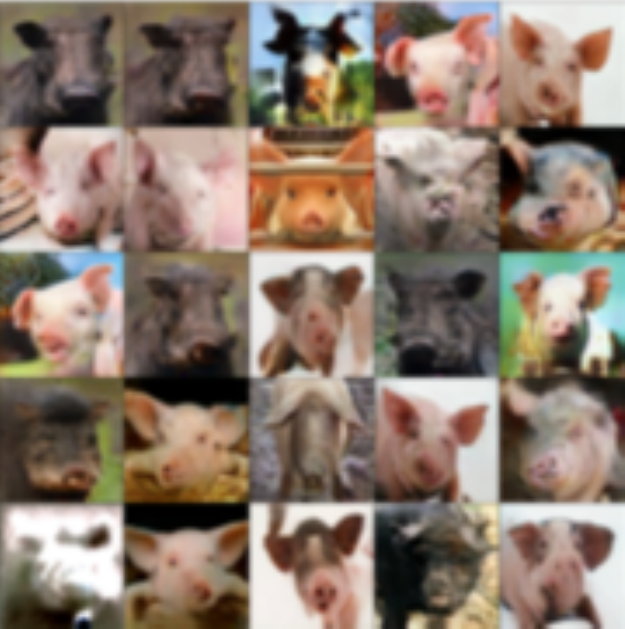}
\caption{Pig}
\end{subfigure}~
\begin{subfigure}{0.32\textwidth}
\includegraphics[width=\textwidth]{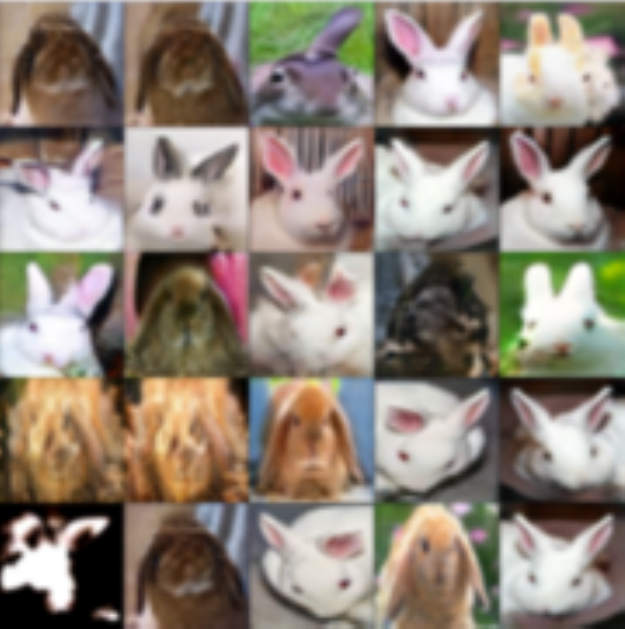}
\caption{Rabbit}
\end{subfigure}
\end{figure}

\clearpage

\begin{figure}[H]
\centering
\ContinuedFloat 
\begin{subfigure}{0.32\textwidth}
\includegraphics[width=\textwidth]{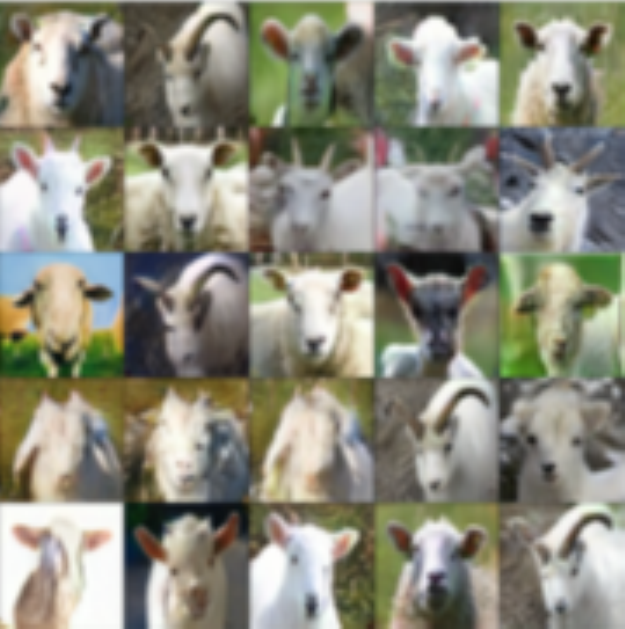}
\caption{Sheep}
\end{subfigure}~
\begin{subfigure}{0.32\textwidth}
\includegraphics[width=\textwidth]{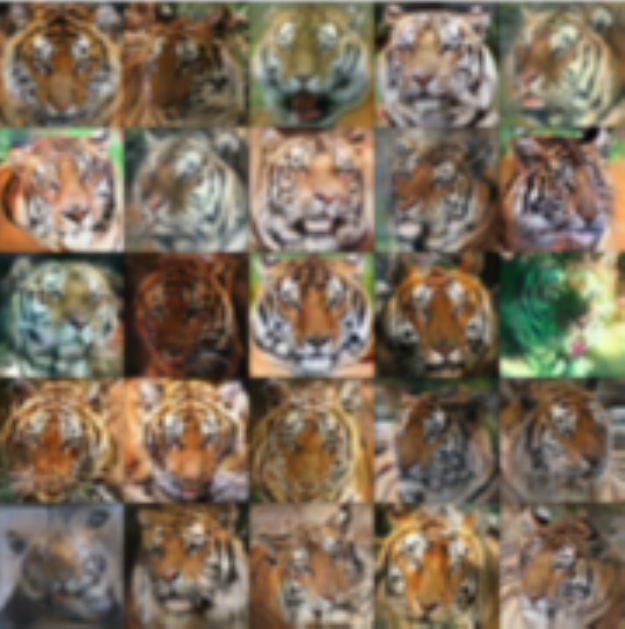}
\caption{Tiger}
\end{subfigure}~
\begin{subfigure}{0.32\textwidth}
\includegraphics[width=\textwidth]{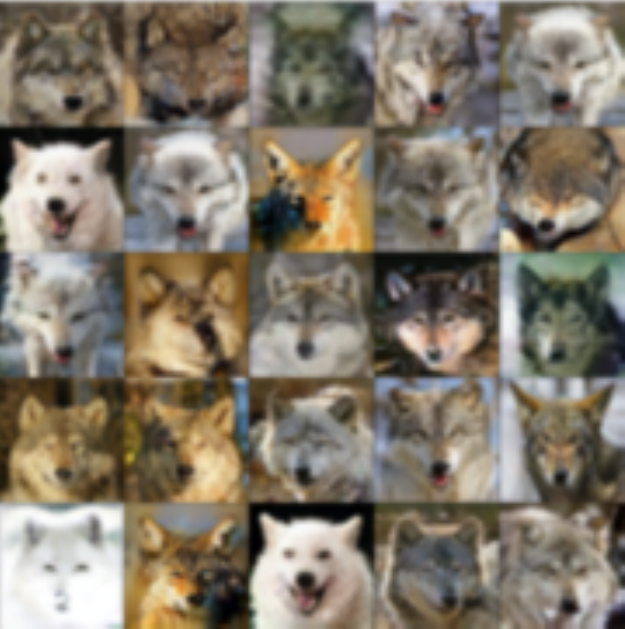}
\caption{Wolf}
\end{subfigure}
\end{figure}

\begin{figure}[H]
\centering
\ContinuedFloat 
\begin{subfigure}{0.32\textwidth}
\includegraphics[width=\textwidth]{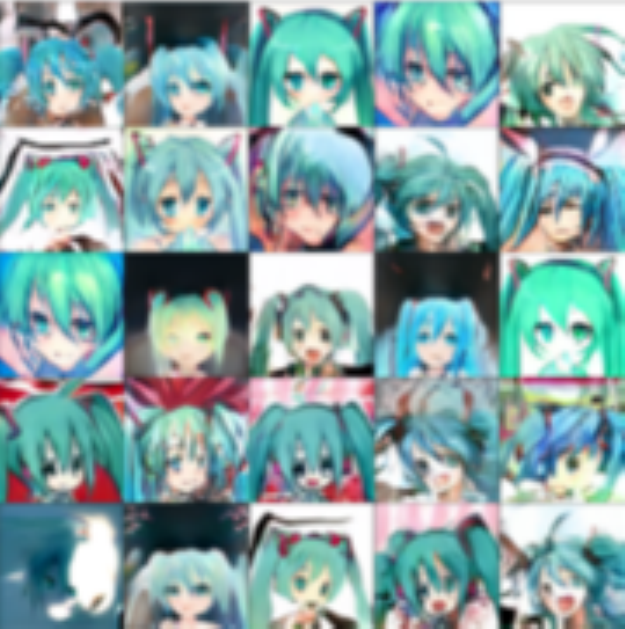}
\caption{Miku}
\end{subfigure}~
\begin{subfigure}{0.32\textwidth}
\includegraphics[width=\textwidth]{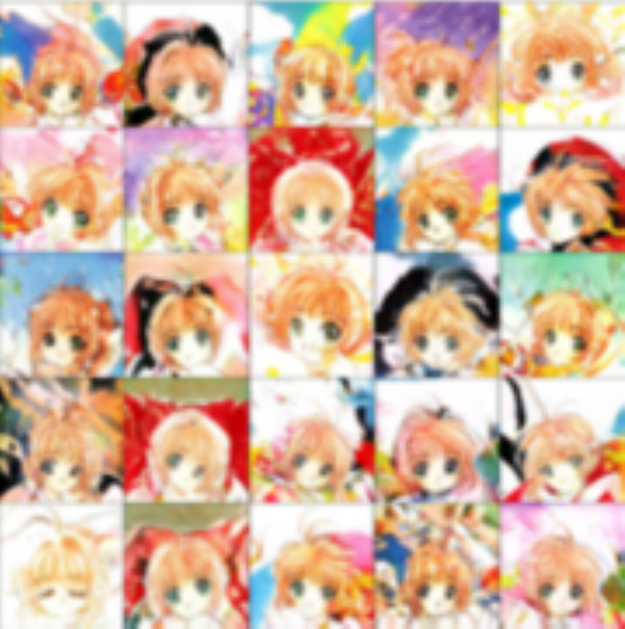}
\caption{Sakura}
\end{subfigure}~
\begin{subfigure}{0.32\textwidth}
\includegraphics[width=\textwidth]{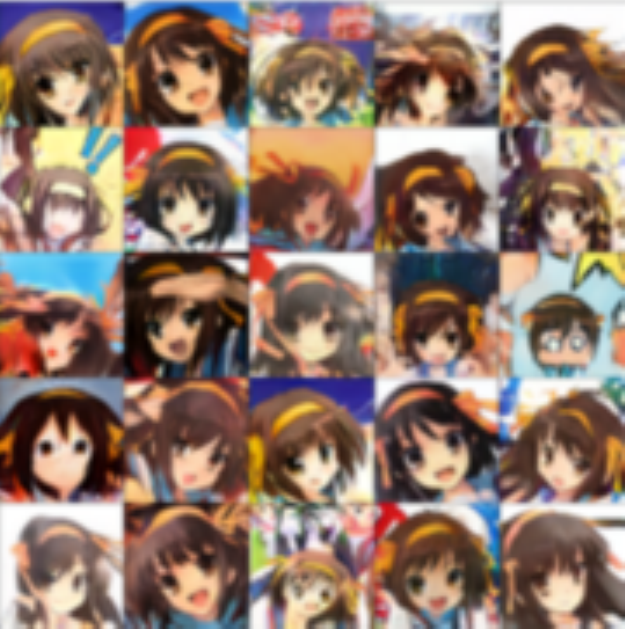}
\caption{Haruhi}
\end{subfigure}
\end{figure}

\begin{figure}[H]
\centering
\ContinuedFloat 
\begin{subfigure}{0.32\textwidth}
\includegraphics[width=\textwidth]{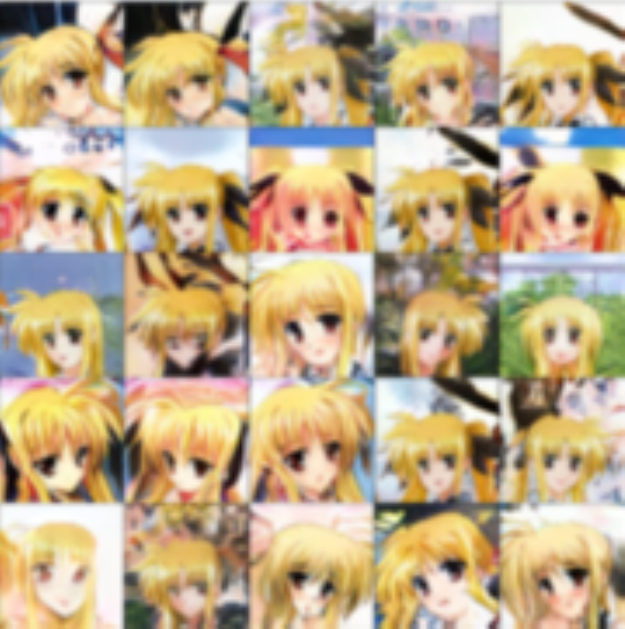}
\caption{Fate}
\end{subfigure}~
\begin{subfigure}{0.32\textwidth}
\includegraphics[width=\textwidth]{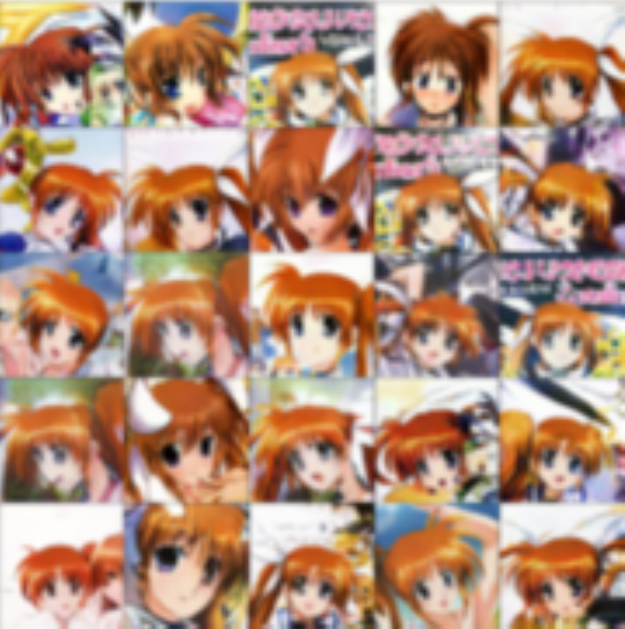}
\caption{Nonoha}
\end{subfigure}~
\begin{subfigure}{0.32\textwidth}
\includegraphics[width=\textwidth]{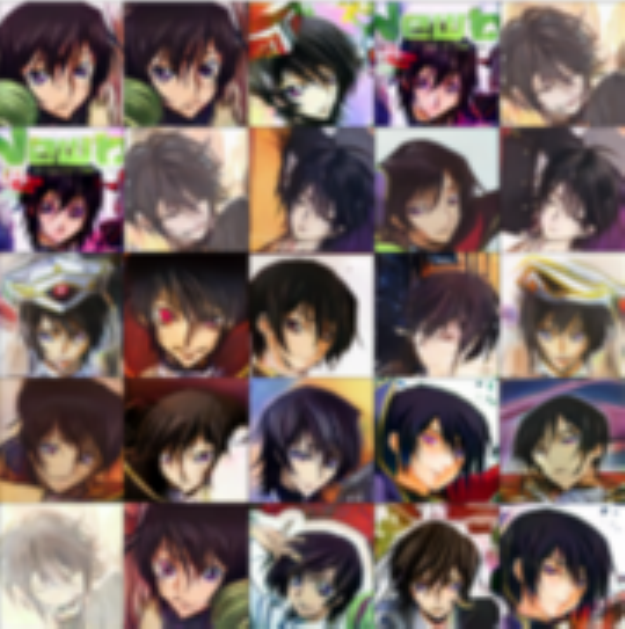}
\caption{Lelouch}
\end{subfigure}
\end{figure}

\clearpage

\begin{figure}[H]
\centering
\ContinuedFloat 
\begin{subfigure}{0.32\textwidth}
\includegraphics[width=\textwidth]{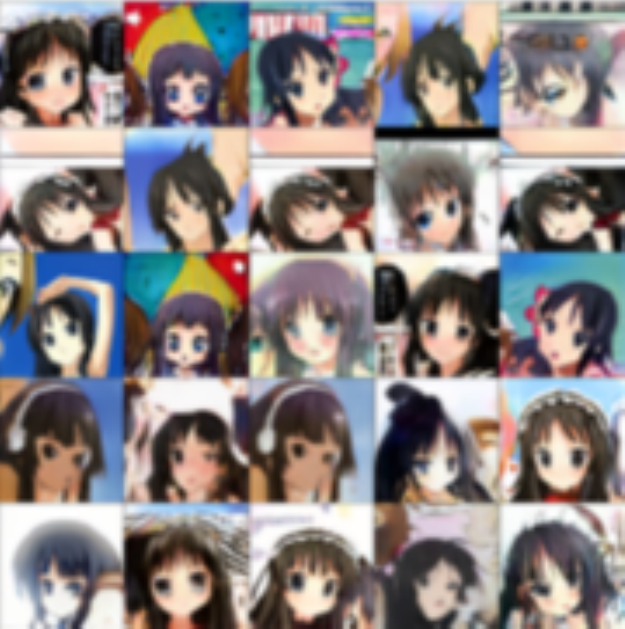}
\caption{Mio}
\end{subfigure}~
\begin{subfigure}{0.32\textwidth}
\includegraphics[width=\textwidth]{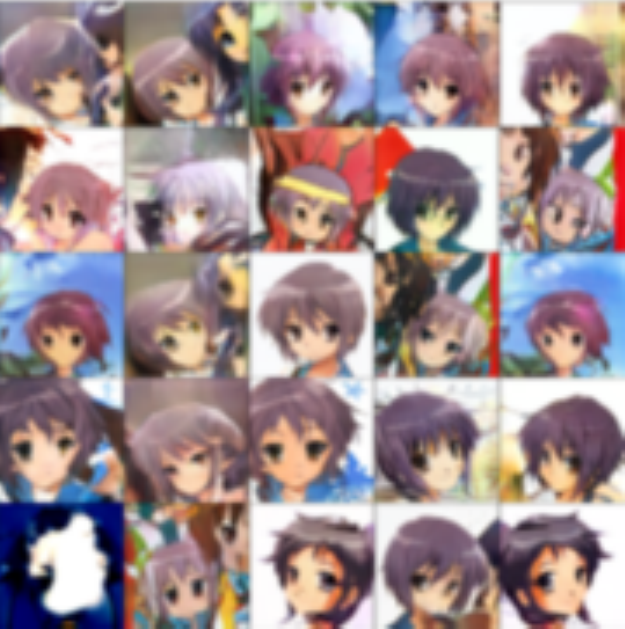}
\caption{Yuki}
\end{subfigure}~
\begin{subfigure}{0.32\textwidth}
\includegraphics[width=\textwidth]{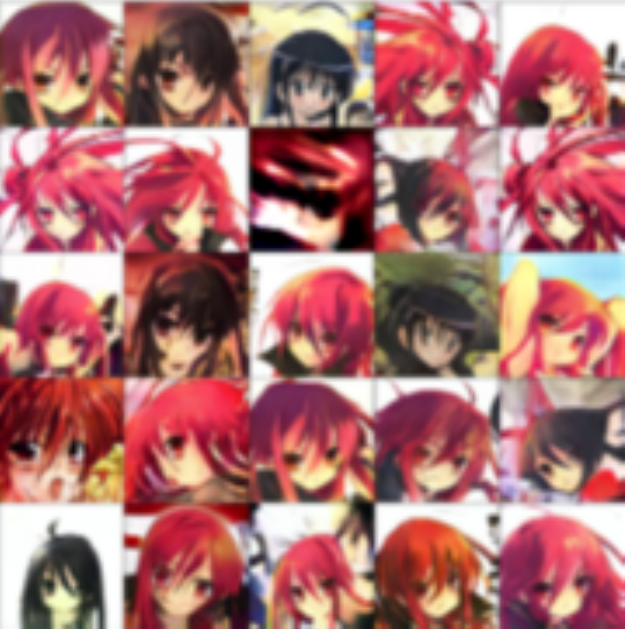}
\caption{Shana}
\end{subfigure}
\end{figure}

\begin{figure}[H]
\centering
\ContinuedFloat 
\begin{subfigure}{0.32\textwidth}
\includegraphics[width=\textwidth]{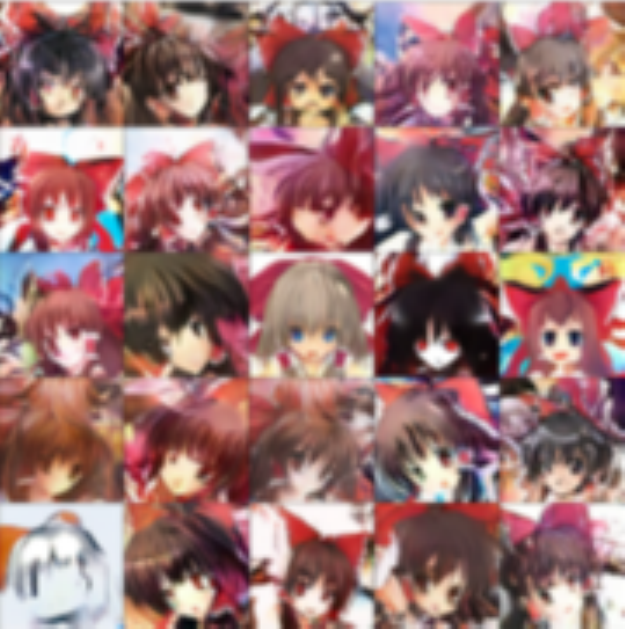}
\caption{Reimu}
\end{subfigure}~
\begin{subfigure}{0.32\textwidth}
~
\end{subfigure}~
\begin{subfigure}{0.32\textwidth}
~
\end{subfigure}
\end{figure}

\clearpage
\section{Generated Samples by SNGAN-projection}
\label{sec:a_samples_projection}

\begin{figure}[H]
\centering
\begin{subfigure}{0.49\textwidth}
\includegraphics[width=\textwidth]{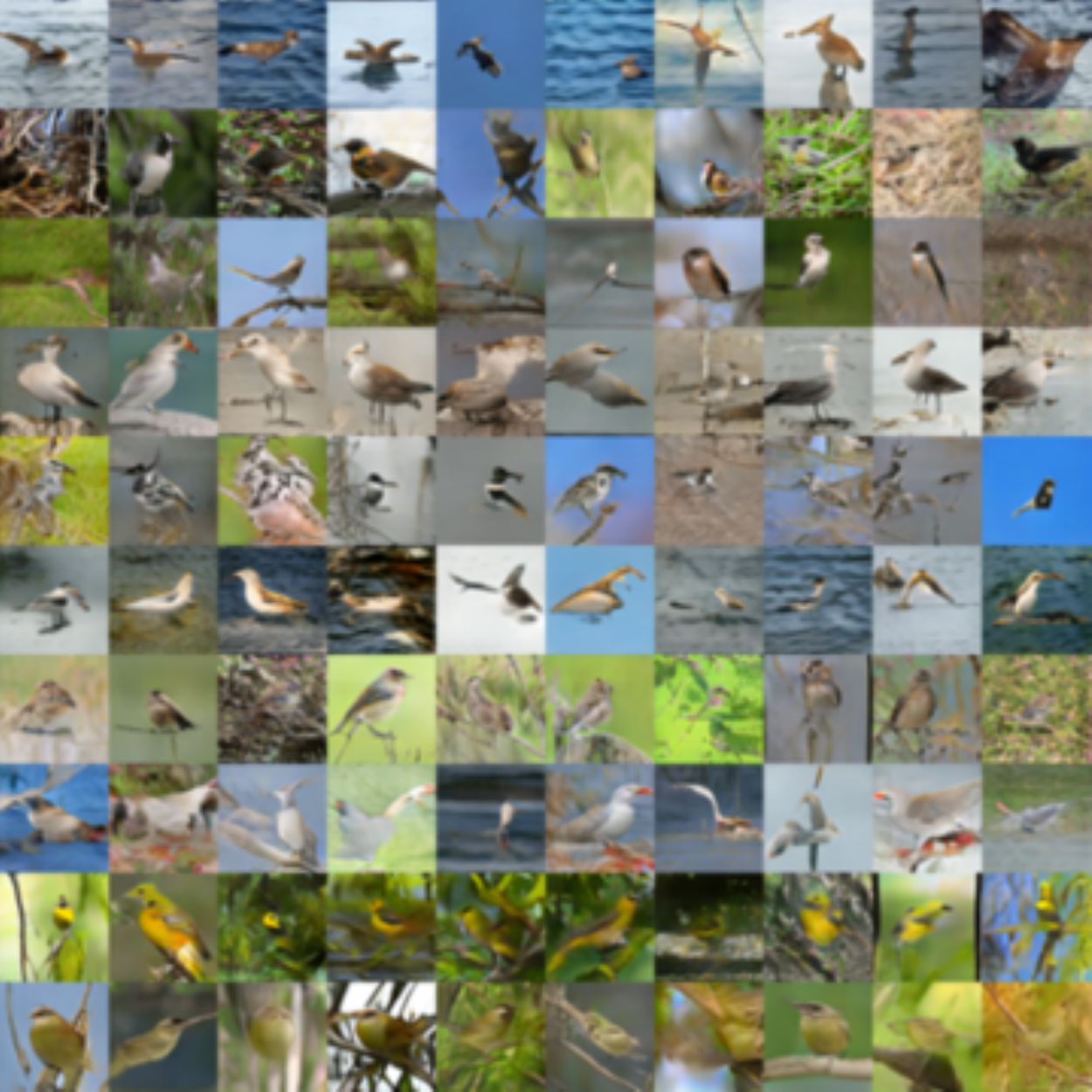}
\caption{CUB-200-2011 (fine-tuning)}
\end{subfigure}~
\begin{subfigure}{0.49\textwidth}
\includegraphics[width=\textwidth]{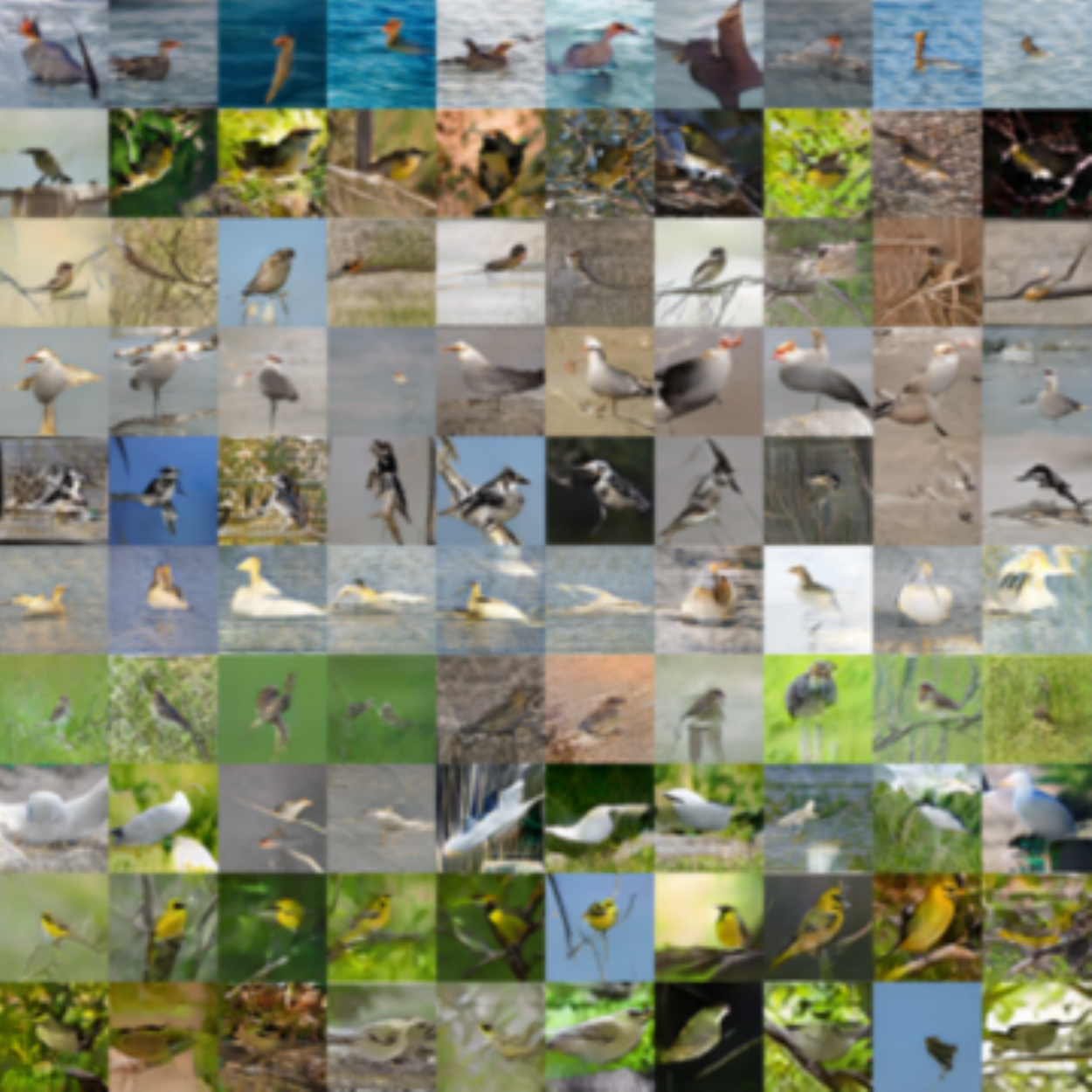}
\caption{CUB-200-2011 (FreezeD)}
\end{subfigure}~
\end{figure}

\begin{figure}[H]
\centering
\ContinuedFloat 
\begin{subfigure}{0.49\textwidth}
\includegraphics[width=\textwidth]{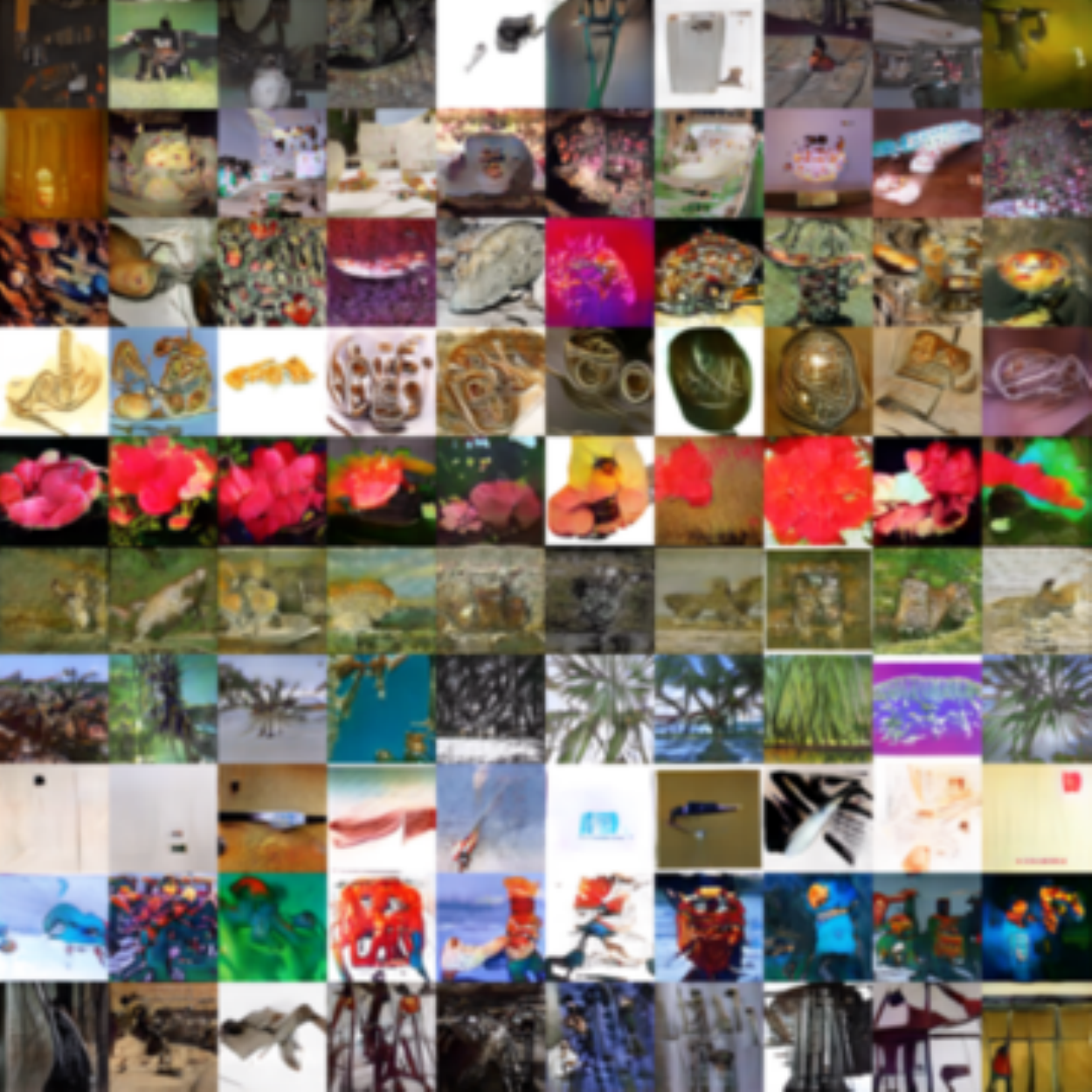}
\caption{Caltech-256 (fine-tuning)}
\end{subfigure}~
\begin{subfigure}{0.49\textwidth}
\includegraphics[width=\textwidth]{figures/caltech_base.pdf}
\caption{Caltech-256 (FreezeD)}
\end{subfigure}~
\end{figure}

\end{document}